\documentclass{article}

\usepackage[preprint]{neurips_2024}

\usepackage[utf8]{inputenc} 
\usepackage[T1]{fontenc}    
\usepackage{hyperref}       
\usepackage{url}            
\usepackage{booktabs}       
\usepackage{amsfonts}       
\usepackage{nicefrac}       
\usepackage{microtype}      
\usepackage{amsmath} 
\usepackage[table]{xcolor}
\usepackage{multicol}
\usepackage{multirow}
\usepackage{graphicx}
\usepackage{wrapfig}
\usepackage[english]{babel}
\usepackage{amsthm}
\newtheorem{proposition}{Proposition}[section]

\usepackage{natbib}
\setcitestyle{numbers,square}

\title{QUAD: Quantization and Parameter-Efficient Tuning of LLM with Activation Decomposition}

\author{
  Yuxuan Hu \\
  Renmin University of China\\
  \texttt{huyuxuan1999@ruc.edu.cn} \\
  \And
  Xiaodong Chen \\
  Renmin University of China\\
  \texttt{chenxiaodong@ruc.edu.cn}
  \And
  Cuiping Li\\
  Renmin University of China\\
  \texttt{licuiping@ruc.edu.cn}\\
  \And
  Hong Chen\\
  Renmin University of China\\
  \texttt{chong@ruc.edu.cn} \\
  \And
  Jing Zhang \\
  Renmin University of China\\
  \texttt{zhang-jing@ruc.edu.cn}\\
}

\newcommand{\vpara}[1]{\vspace{1.5ex}\noindent\textbf{#1}}

\begin{document}

\maketitle

\begin{abstract}
Large Language Models (LLMs) excel in diverse applications but suffer inefficiency due to massive scale. While quantization reduces computational costs, existing methods degrade accuracy in medium-sized LLMs (e.g., Llama-3-8B) due to activation outliers. To address this, we propose QUAD (Quantization with Activation Decomposition), a framework leveraging Singular Value Decomposition (SVD) to suppress activation outliers for effective 4-bit quantization. QUAD estimates activation singular vectors offline using calibration data to construct an orthogonal transformation matrix P, shifting outliers to additional dimensions in full precision while quantizing rest components to 4-bit. Additionally, QUAD enables parameter-efficient fine-tuning via adaptable full-precision outlier weights, narrowing the accuracy gap between quantized and full-precision models. Experiments demonstrate that QUAD achieves 94–96\% accuracy under W4A4 quantization and 98\% accuracy with W4A4/A8 and parameter-efficient fine-tuning for Llama-3 and Qwen-2.5 models. Our code is available at \href{https://github.com/hyx1999/Quad}{repository}.
\end{abstract}

\section{Introduction}

Large Language Models (LLMs)~\cite{llama3-grattafiori2024, qwen2-yang2024, deepseek-liu2024} have demonstrated remarkable performance across numerous fields and have been widely adopted in various applications, such as chat assistants, coding copilots~\cite{code-llama-roziere2023, qwen2-coder-hui2024, codes-haoyangli2024}, and beyond. However, the scaling law~\cite{scalinglaw-kaplan2020} has led to increasingly deeper LLMs with hundreds of billions of parameters, rendering them inefficient for text-processing tasks. Concurrently, existing work~\cite{nanoflow-zhu2024} highlights that for throughput-oriented serving systems, many workloads are predominantly compute-bound. Consequently, there is a critical need for effective methods to compress LLMs and enhance the efficiency of General Matrix Multiplication (GEMM) operations, given that GEMM dominates computational tasks within LLMs.

Weight and activation quantization aims to address this issue by representing both weights and activations in lower precision, thereby reducing storage overhead and leveraging more efficient hardware, such as INT4 tensor cores, to accelerate GEMM computations. While existing approaches~\cite{gptq-frantar2022, awq-lin2024, bitdistiller-du2024} have successfully quantized LLM weights to 4 bits or even lower with nearly no loss in accuracy, quantizing both weights and activations remains challenging. This difficulty arises due to the higher prevalence of outliers in activations, making them harder to quantify than weights. Existing methods like QuaRot~\cite{quarot-ashkboos2024}, SpinQuant~\cite{spinquant-liu2024}, and DuQuant~\cite{duquant-lin2024} attempt to suppress outliers in both weights and activations by rotating the matrix and quantizing both to 4 bits. These techniques perform well for large-scale LLMs, such as Llama-2-70B, yet still result in significant performance degradation for medium-sized LLMs, such as Llama-3-8B. We attribute this to the fact that for medium-sized LLMs, the rotation matrix is insufficient in suppressing outliers, making it difficult to quantize activations to 4 bits. Thus, more effective methods are required to eliminate outliers from activations.

Singular Value Decomposition (SVD) is an effective tool used to decompose a matrix's high and low-frequency components, thus enabling the removal of outliers from the matrix~\cite{svdqunat-li2024}. Leveraging the impressive performance of SVD in eliminating outliers, we apply it to activations and propose our method, named \textbf{QUAD} (\textbf{Qu}antization with \textbf{A}ctivation \textbf{D}ecomposition). However, migrating SVD to activations presents several challenges. \textbf{Firstly}, weights remain static, whereas activations dynamically change with varying inputs during serving. \textbf{Secondly}, while the SVD of the weight matrix can be computed offline, the SVD of activations cannot be performed online. \textbf{Thirdly}, the introduced SVD must be compatible with existing rotation methods. To address these challenges, QUAD estimates the singular vectors of activations~\cite{sp3-hu-etal-2024, slicegpt-ashkboos2024} offline using a small amount of calibration data and constructs the transformation matrix \({ P} \in \mathbb{R}^{(C+r) \times C}\) based on these singular vectors. The matrix \( P\) has two key properties: (1) it shifts outliers in activations to an additional \(r\) dimension, thereby eliminating outliers in the original activations; (2) it satisfies \( P^\top P = I\). Consequently, matrix \( P\) enables equivalent transformations of the LLM and is compatible with existing methods. After transformation, for GEMM computations, most weights and activations are quantized to 4 bits, while the \(r\) outlier dimensions and their corresponding weights are retained in full precision.

Beyond removing activation outliers, QUAD can also be applied to parameter-efficient fine-tuning of quantized models~\cite{lora-hu2022, qlora-dettmers2023, rolora-huang2024}. Specifically, we retain the weights corresponding to the outlier dimensions in full precision, meaning these portions of weights can serve as parameter-efficient adapters to fine-tune the model. Additionally, we demonstrate that the adapter initialized by QUAD provides a sub-optimal solution for approximating full fine-tuning with unchanged input distributions. Based on the full-precision portion of the model, we further reduce the gap between the quantized and full-precision models through parameter-efficient fine-tuning.

To evaluate the effectiveness of QUAD, we conducted extensive experiments on diverse LLMs and datasets. The contributions of this work are summarized as follows:

\begin{itemize}
    \item We introduce QUAD, which suppresses outliers in activations via SVD decomposition and integrates with existing rotation methods to enhance the performance of quantized models.
    \item Beyond quantization, QUAD can also be used for parameter-efficient fine-tuning, and we propose fine-tuning the full-precision portion of the quantized model to further bridge the gap between the quantized and full-precision models.
    \item Based on these improvements, QUAD maintains 94\% to 96\% of the full-accuracy model's performance under W4A4 quantization and achieves 98\% of the full-accuracy model's performance when combined with W4A4/A8 quantization and parameter-efficient fine-tuning.
\end{itemize}

\section{Related Work}

Quantization is a commonly used approach for LLM deployment to compress storage space and improve inference speed by representing weights and activations in lower bits. Existing quantization methods can be broadly categorized into two categories: Quantization-aware training (QAT) and Post-training quantization (PTQ). QAT combines quantization with training and fine-tuning to represent the model with lower precision while maintaining model performance. Representative QAT methods include LLM-QAT~\cite{llmqat-liu2023}, BitDistiller~\cite{bitdistiller-du2024}, EfficientQAT~\cite{efficientqat-chen2024}, and the BitNet series~\cite{bitnet-wang2023, bitneta4.8-wang2024}. Since the QAT method has large resources and time overhead, more work focuses on PTQ, which can achieve quantization with only a small amount of calibration data. The main challenge of PTQ methods comes from the outliers in the parameters and activations of LLM, which bring large quantization errors. Therefore, existing methods have proposed the following methods to overcome the impact of outliers, including model equivalent transformation and weight compensation. Model equivalent transformation, including shifting, scaling, and rotation. SmoothQuant~\cite{smoothquant-xiao2023} and AWQ~\cite{awq-lin2024} propose employing scaling operations, while OS+~\cite{outlier-wei2022, outlier-plus-wei2023} proposes shifting operations to suppress outliers. In addition to scaling and shifting, QUIP~\cite{quip-chee2023}, QuaRot~\cite{quarot-ashkboos2024}, DuQuant~\cite{duquant-lin2024}, and SpinQuant~\cite{spinquant-liu2024} further utilize rotation operations to suppress outliers. These transformations are also used in subsequent work, such as QUIK~\cite{quik-ashkboos2023}, QmniQuant~\cite{omniquant-shao2023}, AffineQuant~\cite{affinequant-ma2024}, QServe~\cite{qserve-lin2024}, and OSTQuant~\cite{ostquant-hu2025}. The weight compensation technique, which improves quantization by adjusting the weights during the quantization process, was first proposed by OBS~\cite{OBS-hassibi1993} and subsequently widely applied to LLM by GPTQ~\cite{gptq-frantar2022}. In addition to transformations and compensation, existing work also considered the use of mixed-precision~\cite{llm.int8-dettmers2022} and non-uniform data types~\cite{gptvq-van2024, vptq-liu2024} to improve quantization.

\section{Background}


\begin{figure}[t]
    \begin{minipage}[t]{0.6\textwidth}
        \centering
        \includegraphics[width=\linewidth]{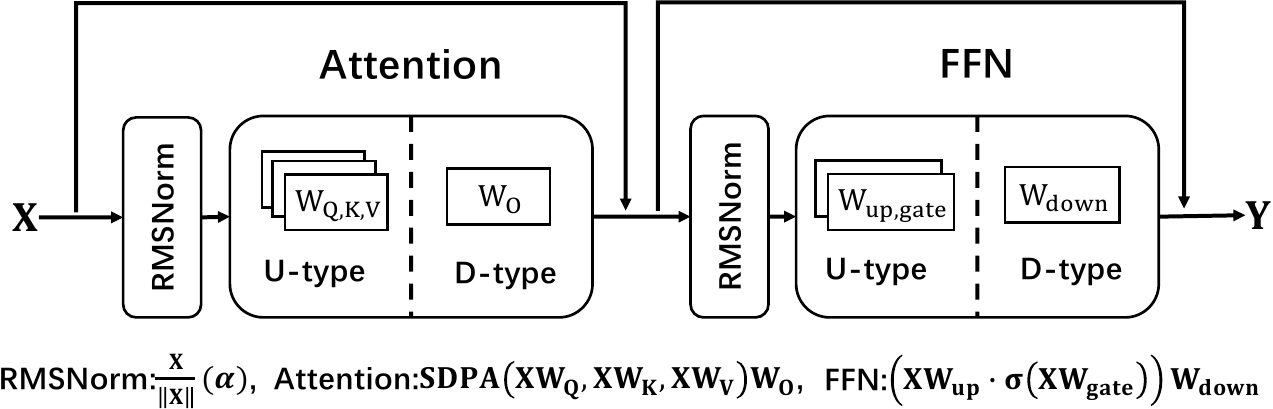} 
        \caption{Example of a transformer layer structure.}
        \label{fig: structure}
    \end{minipage}
    \begin{minipage}[t]{0.4\textwidth}
      \centering
      \includegraphics[width=\linewidth]{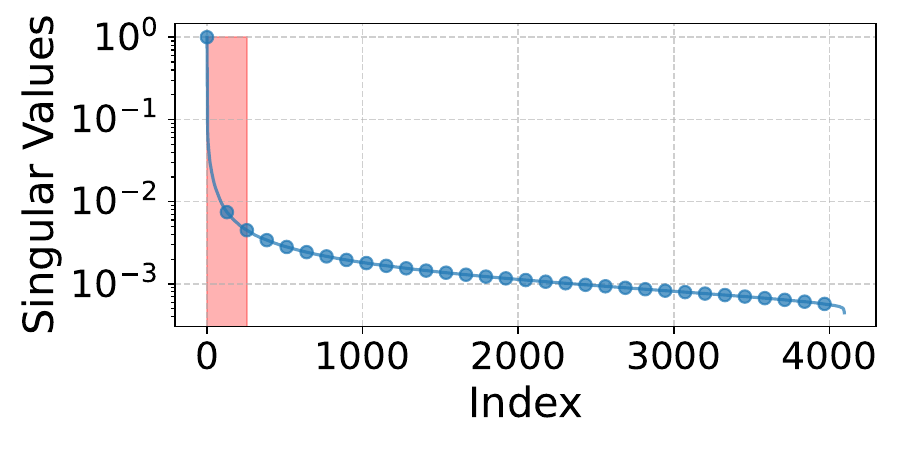} 
      \caption{Estimated singular values.}
      \label{fig: singular values}
    \end{minipage}
\end{figure}

\subsection{Transformer Architecture}

A large language model (LLM) typically comprises an embedding layer, a sequence of transformer layers, and a language model head. Figure~\ref{fig: structure} depicts the architecture of a standard transformer layer, which consists of three core components: the RMSNorm module, the Attention module, and the Feed-Forward Network (FFN) module. These layers collectively involve seven weight matrices: $W_Q$, $W_K$, $W_V$, $W_O$, $W_{\text{up}}$, $W_{\text{gate}}$, and $W_{\text{down}}$. To analyze their roles systematically, we categorize these matrices into two groups based on their position relative to module inputs and outputs. Specifically, \textbf{U-type matrices}—those positioned near module inputs—include $W_Q$, $W_K$, $W_V$, $W_{\text{up}}$, and $W_{\text{gate}}$, while \textbf{D-type matrices}—situated near outputs—comprise $W_O$ and $W_{\text{down}}$.

\subsection{Equivalence Transformations}

Prior research demonstrates that an LLM's output remains invariant under certain linear transformations of its weight matrices. A key technique involves applying an orthogonal matrix $Q$ ($Q^\top Q = I$) to perform equivalence transformations. For U-type matrices, this entails left-multiplying by $Q$, followed by right-multiplying their corresponding D-type matrices by $Q^\top$ to preserve the overall computation. This approach remains valid even in the presence of RMSNorm layers between modules, as shown by the following identity:

\begin{equation}
    \text{RMSNorm}(X) = \text{RMSNorm}(X Q^\top) Q.
\end{equation}

Practically, we first absorb the scaling parameters of adjacent RMSNorm modules into neighboring weight matrices as described in prior work like QuaRot. After this absorption, the RMSNorm operation simplifies to $\text{RMSNorm}(x) = x / \lVert x \rVert$. Then, we can transform the weights by rotating the matrix, \(W_U \leftarrow Q W_U\), \(W_D \leftarrow W_D Q^\top\), where \(W_U\), \(W_D\) denotes the U-type and D-type weight matrices. It should be noted that the equivalence transformation is not limited to rotation matrices; any matrix $Q$ (satisfying $Q^\top Q = I$) can be used to transform the weights of the model while preserving its output unchanged.

\section{Method}

Our proposed method, QUAD, comprises three stages: transformation (Section~\ref{section: transformation}), quantization (Section~\ref{section: quantization}), and parameter-efficient tuning (Section~\ref{section: tuning}). In the transformation stage, we utilize singular vectors of activations to construct a projection matrix for mapping outliers in the activations to additional dimensions. Subsequently, weight matrices and activations are smoothed using a rotation matrix. Following the approach of QuaRot, in the quantization phase, we apply GPTQ to quantize the weight matrix and inject quantization operators into the model to quantify activations online using the round-to-nearest method. Notably, the weights and activations corresponding to the outlier dimensions retain full precision. Finally, in the parameter-efficient tuning stage, we fine-tune the full-precision part of the quantized model using high-quality data.

\subsection{Transformation\label{section: transformation}}

Model equivalence transformation aims to project outliers in activations into additional dimensions and smooth the original weight matrices and activations. To achieve this, we first fuse the scaling parameters \((\alpha)\) of each RMSNorm module into adjacent weight matrices. Next, we construct a projection matrix based on calibration data and singular value decomposition (SVD). Let \({ X}^{(i)} \in \mathbb{R}^{B \times h}\) denote the input activation of layer \(i\) of the model, where \(B\) represents the number of tokens in the calibration data, and \(h\) represents the dimensions of the activation. We can estimate the singular vectors of the activation using the following formula:

\begin{equation}
    { U, \Sigma, U^\top} = \text{SVD}(\sum_{i} { X^{(i)\top} X^{(i)}}).
\end{equation}

Here, the columns of \({ U}\) correspond to the estimated singular vectors of the activation, and \({\Sigma}\) contains the corresponding singular values. Figure~\ref{fig: singular values} shows the magnitude of the singular values for different singular vectors, revealing that a small subset of singular vectors dominates the singular values. We hypothesize that these dominant singular vectors contribute to the presence of outliers in the activation. To mitigate this, we propose removing the components associated with these dominant singular vectors from the existing activations and projecting them into additional dimensions. Specifically, if the first \(r\) singular vectors are to be removed, the projection matrix can be constructed as follows:

\begin{equation}
    { P} = 
    \left(
    \begin{matrix}
        { U}_{1:r}, { I} - \sum_{i=1}^{r} { U}_i { U}_i^\top
    \end{matrix}
    \right) \in \mathbb{R}^{h \times (r + h)}
    \label{formula: projection}
\end{equation}

It can be proven that the matrix \( P\) satisfies \( PP^\top = I\) (see Appendix~\ref{section: proof of PPt = I}), making it suitable for model equivalence transformation. Furthermore, by letting \( \hat{X} = XP\), the outliers in \( X\) can be projected into the first \(r\) dimensions of \( \hat{X}\). Subsequently, we employ a random Hadamard matrix \( H \) to construct a rotation matrix \( Q \), which further smooths the weights and activations.

\begin{equation}
    Q = \left(
    \begin{matrix}
        I_{r \times r} & 0 \\
        0 & H_{h \times h}
    \end{matrix} \right) \in \mathbb{R}^{(r+h) \times (r+h)}.
\end{equation}

\noindent Here, the Hadamard matrix is a specialized rotation matrix that can be utilized to smooth weights and activations and can be computed efficiently. Please refer to Appendix~\ref{section: hadamard} for a more detailed explanation of the Hadamard matrix.

Finally, the introduced projection and rotation matrices are fused with the existing weight matrices. For a U-type weight matrix \(W_U\) and a D-type weight matrix \(W_D\), they are updated as follows:
\[ W_U \leftarrow Q^\top P^\top (\alpha) W_U, \quad W_D \leftarrow W_D { P} { Q}. \]

\begin{figure}[t]
    \centering
    \includegraphics[width=\linewidth]{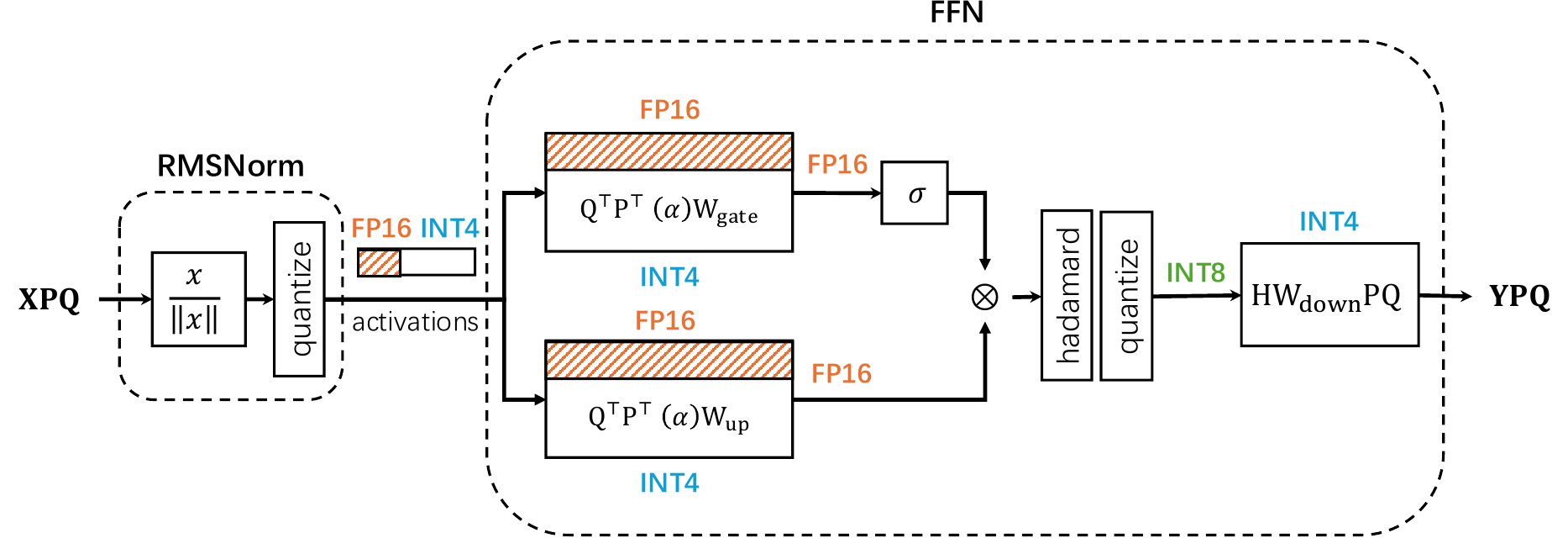} 
    \caption{The FFN module of the transformer layer after applying QUAD transformation.}
    \label{fig: ffn quad}
\end{figure}

Figure~\ref{fig: ffn quad} illustrates the FFN module of the transformed model and the Attention module of the transformed model is shown in Appendix~\ref{section: quad attention}. By combining the projection and rotation, we smooth the matrix and activations and project the outliers into additional \(r\) dimensions. Additionally, following QuaRot and SpinQuant, we add online Hadamard operators before \(W_O\) and \(W_{down}\) to smooth the intermediate activations of the Attention module and the FFN module. Meanwhile, for a D-type matrix \(W_D\), we further multiply the Hadamard matrix \(H\) on its left-hand side:
\[ W_D \leftarrow { H} W_D { P} { Q}. \]

In our experiment, we additionally found that for models where the number of attention heads is not a power of 2, the online Hadamard transform leads to high latency, so for such models, we explore how to eliminate the online Hadamard transform in Appendix~\ref{section: low-rank branch}.

\begin{figure}[t]
    \centering
    \includegraphics[width=\linewidth]{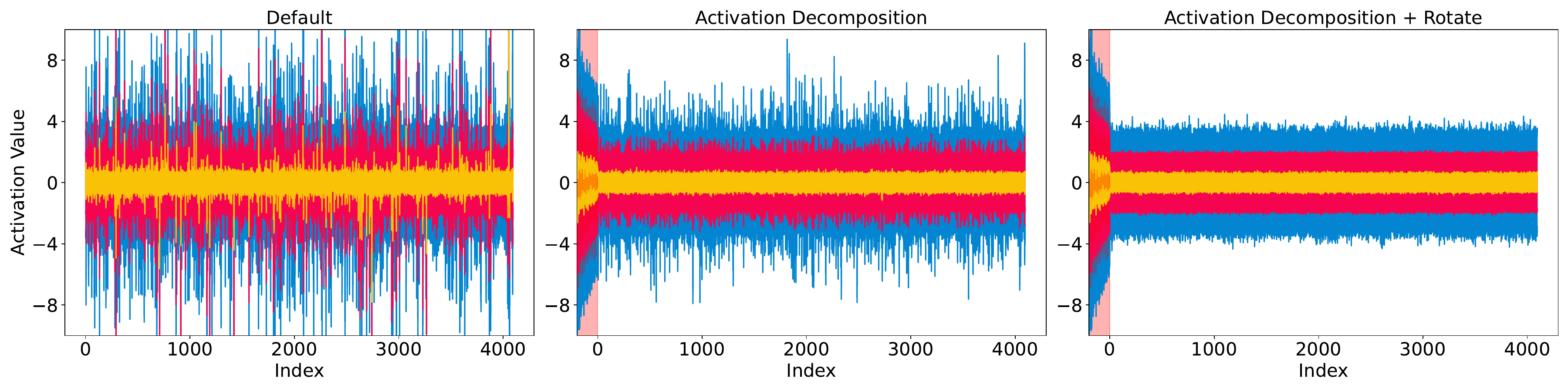} 
    \caption{The input activations of the transformer layer after applying QUAD transformation.}
    \label{fig: activations}
\end{figure}

\subsection{Quantization\label{section: quantization}}

To enhance the computational efficiency of GEMM using INT4 and INT8 TensorCore, we quantize both weights and activations using a symmetric per-row (per-token) approach. Specifically, after applying transformations, we use GPTQ to quantize the weights of large language models (LLMs). Before each linear layer, we quantize activations online using the round-to-nearest (RTN) method. This process involves dividing the maximum absolute value of each token by the highest value expressible with the target quantization precision (7 for INT4 and 127 for INT8) to determine the scale for each token. Each token is then divided by its corresponding scale and rounded to the nearest integer.

Furthermore, we introduce an extra \(r\) dimension for activations preceding U-type linear layers to represent outliers. This ensures that the weights and activations corresponding to this part remain at full precision during quantization while the remaining components are quantized using INT4. Previous studies have noted distributional differences between activations before U-type and D-type linear layers, observing that activations before D-type layers are more challenging to quantify. Consequently, for activations preceding D-type linear layers, we use INT8 for their quantization, whereas the associated weight matrices are quantized using INT4. 

\subsection{Quantization-Aware Parameter-Efficient Tuning\label{section: tuning}}

\begin{wrapfigure}{l}{0.5\textwidth}
    \centering
    \includegraphics[width=\linewidth]{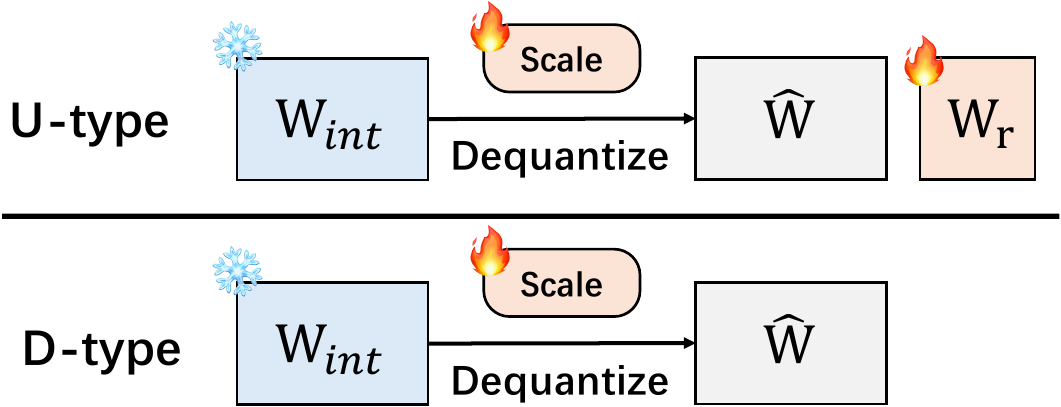} 
    \caption{End-to-end quantization-aware tuning.}
    \label{fig: tuning}
\end{wrapfigure}

The additional weights corresponding to outliers can also be leveraged for efficient model parameter tuning. Thus, our goal is to minimize the gap between the quantized and full-precision models through fine-tuning with high-quality datasets. As shown in Figure~\ref{fig: tuning}, refinement is applied to the full-precision sub-matrix (\(W_r\)) associated with outlier dimensions for U-type layers. Meanwhile, following EfficientQAT~\cite{efficientqat-chen2024}, we also adjust the scale corresponding to the quantization matrix for both U-type and D-type weight matrices. We adopt the straight-through estimator~\cite{STE} (STE) for gradient transmission to handle non-differentiable operations, such as quantization during model training, i.e., \( \nabla X \leftarrow \nabla X_q \), where \(X_q = \text{Dequantize}(\text{Quantize}(X))\).

\subsection{Theoretical Analyses}

\vpara{Quantization}

Assume the input to a linear layer is \(X \in \mathbb{R}^{B \times m}\) and the weight matrix is \(W \in \mathbb{R}^{m \times n}\). Following SVDQuant~\cite{svdqunat-li2024}, we define the quantization error as \(E(X, W) = \Vert XW - Q(X)Q(W) \Vert_F\), where \(\Vert \cdot \Vert_F\) denotes the Frobenius norm, and further present the following propositions:

\begin{proposition}
    The quantization error can be decomposed as follows:
    \[
        E(X, W) \leq \Vert X \Vert_F \Vert W - Q(W) \Vert_F + \Vert X - Q(X) \Vert_F \Vert Q(W) \Vert_F.
    \]\label{prop: quant error}
\end{proposition}

\begin{proposition}
    For the round-to-nearest (RTN) quantization method \(Q\), if the tensor \(X\) follows a normal distribution, we have:
    \[
        \mathbb{E} \left[ \Vert X - Q(X) \Vert_F \right] \leq \frac{\sqrt{\log (size(X) \pi)}}{q_{\text{max}}} \mathbb{E} \left[ \Vert X \Vert_F \right],
    \]\label{prop: rtn bound}
\end{proposition}

Our core idea is to reduce the difficulty of quantifying the input by projecting outliers to additional dimensions. Specifically, we decompose the inputs and weights into two parts: \(\hat{X}, \hat{W}\) and \(X_r, W_r\), respectively, such that \(XW = \hat{X}\hat{W} + X_rW_r\). Therefore, we can rewrite the quantization error as follows:

\[
    E(X, W) = \Vert XW - X_rW_r - Q(\hat{X})Q(\hat{W}) \Vert_F = \Vert \hat{X}\hat{W} - Q(\hat{X})Q(\hat{W}) \Vert_F.
\]

From the propositions above, it is evident that the quantization error correlates with both the magnitude of the inputs and their quantization errors. Moreover, the quantization error is further unified with the input magnitude, as the quantization error of the input is inherently limited by its magnitude. Consequently, reducing the input \(X\)'s magnitude emerges as a viable strategy to minimize the quantization error. Since \(\Vert X \Vert_F = \sqrt{\sum_{i=1}^{\min(n,m)} \sigma_i^2}\), our approach involves minimizing the input's norm by eliminating the largest \(r\) singular values. This process is formalized as \(X_r = X \left(U_{1:r} \right)\) and \(\hat{X} = X \left(I - \sum_{i=1}^{r}U_i U_i^\top\right)\), where \(U,\Sigma, U^\top = \text{SVD}(X^\top X)\).

\vpara{Parameter-Efficient Fine-Tuning}

In addition to reducing the quantization error, the decomposed weight matrix \(W_r\) can also be used for parameter-efficient fine-tuning of the quantized model.

\begin{proposition}
    Parameter-efficient tuning via \(W_r = U_{1:r}^\top W\) is a suboptimal solution to approximate full fine-tuning with unchanged input distribution. Specifically, suppose the gradients of the weights \(W\) and \(W_r\) are \(\nabla W\) and \(\nabla W_r\), then we have:
    \[
        X \nabla W \approx X_r \nabla W_r.
    \]\label{prop: approx}
\end{proposition}

Consider the output of the linear layer is \(Y = XW\), and the gradient of \(Y\) is \(\nabla Y\). Given \(\nabla W = X^\top \nabla Y\), then based on the weight gradients and inputs, we can estimate the variation of the linear layer outputs as \(\Delta Y = X \nabla W = X X^\top \nabla Y\). Similarly, we have \(\Delta Y = X_r \nabla W_r = X_r X_r^\top \nabla Y\) for the quantized linear layer. Thus, we can approximate \(XX^\top\) with \(X_r X_r^\top\) such that the parameter-efficient fine-tuning approximates the full fine-tuning, where the optimal solution is \(X_r = X (U_{1:r})\), i.e., \(W_r = (U_{1:r})^\top W\), where \(U,\Sigma, U^\top = \text{SVD}(X^\top X)\). The proof of the above propositions is given in Appendix~\ref{section: proofs}.

\begin{table}[t]
\caption{Zero-shot accuracy of LLAMA-3 models.\label{table: zero-shot llama-3}}
\centering
\renewcommand\arraystretch{1.5}
\resizebox{0.85\linewidth}{!}{
\begin{tabular}{c|c|c|ccccccc}
\hline
\textbf{Model}                & \textbf{Method} & \textbf{Precision} & \textbf{PQ} & \textbf{WG} & \textbf{HS} & \textbf{A-e} & \textbf{A-c} & \textbf{LA} & \textbf{Avg.} \\ \hline
\multirow{5}{*}{Llama-3.2-1B} & Baseline        & FP16               & 74.59       & 60.38       & 63.68       & 60.48        & 36.35        & 62.93       & 59.74         \\
                              & QuaRot          & W4A4               & 67.30       & 54.93       & 53.09       & 52.06        & 31.06        & 43.55       & 50.33         \\
                              & QUAD            & W4A4               & 69.80       & 56.20       & 57.33       & 54.71        & 31.40        & 51.29       & 53.46         \\
                              & QUAD            & W4A4/A8            & 72.80       & 58.72       & 60.71       & 57.15        & 32.85        & 58.12       & 56.73         \\ 
                              & \;+tuning         & W4A4/A8            & 72.80       & 57.77       & 61.70       & 56.52        & 34.39        & 58.84       & 57.00         \\ \hline
\multirow{5}{*}{Llama-3.2-3B} & Baseline        & FP16               & 77.48       & 69.93       & 73.60       & 71.68        & 45.90        & 70.50       & 68.18         \\
                              & QuaRot          & W4A4               & 74.05       & 63.30       & 66.91       & 63.55        & 40.27        & 61.50       & 61.60         \\
                              & QUAD            & W4A4               & 74.65       & 65.75       & 70.01       & 65.11        & 42.41        & 65.48       & 63.90         \\
                              & QUAD            & W4A4/A8            & 75.24       & 65.51       & 71.60       & 67.68        & 43.94        & 68.62       & 65.43         \\ 
                              & \;+tuning         & W4A4/A8            & 76.82       & 67.32       & 72.36       & 69.15        & 44.45        & 67.59       & 66.28         \\ \hline
\multirow{5}{*}{Llama-3-8B}   & Baseline        & FP16               & 80.79       & 73.16       & 79.16       & 77.74        & 53.24        & 75.74       & 73.30         \\
                              & QuaRot          & W4A4               & 75.30       & 66.46       & 72.56       & 67.89        & 42.41        & 67.94       & 65.43         \\
                              & QUAD            & W4A4               & 77.48       & 68.43       & 75.22       & 73.74        & 46.50        & 71.18       & 68.76         \\
                              & QUAD            & W4A4/A8            & 80.09       & 71.35       & 77.12       & 73.82        & 47.53        & 74.35       & 70.71         \\
                              & \;+tuning         & W4A4/A8            & 80.74       & 71.59       & 78.35       & 77.57        & 51.28        & 73.32       & 72.14         \\ \hline
\end{tabular}
}
\end{table}

\begin{table}[t]
\caption{Zero-shot accuracy of Qwen-2.5 models.\label{table: zero-shot qwen2.5}}
\centering
\renewcommand\arraystretch{1.5}
\resizebox{0.85\linewidth}{!}{
\begin{tabular}{c|c|c|ccccccc}
\hline
\textbf{Model}                & \textbf{Method} & \textbf{Precision} & \textbf{PQ} & \textbf{WG} & \textbf{HS} & \textbf{A-e} & \textbf{A-c} & \textbf{LA} & \textbf{Avg.} \\ \hline
\multirow{5}{*}{Qwen2.5-1.5B} & Baseline        & FP16               & 75.95       & 63.46       & 67.73       & 71.68        & 45.48        & 62.06       & 64.39         \\
                              & QuaRot          & W4A4               & 70.13       & 56.35       & 59.78       & 64.06        & 37.03        & 47.51       & 55.81         \\
                              & QUAD            & W4A4               & 72.36       & 59.19       & 62.70       & 68.81        & 41.55        & 56.01       & 60.10         \\
                              & QUAD            & W4A4/A8            & 73.88       & 60.22       & 64.38       & 70.54        & 42.83        & 59.31       & 61.86         \\
                              & \;+tuning       & W4A4/A8            & 74.92       & 60.77       & 65.74       & 69.70        & 41.81        & 59.36       & 62.05         \\  \hline
\multirow{5}{*}{Qwen2.5-3B}   & Baseline        & FP16               & 78.40       & 68.51       & 73.59       & 72.98        & 46.93        & 66.93       & 67.89        \\
                              & QuaRot          & W4A4               & 73.83       & 61.33       & 65.80       & 65.53        & 40.10        & 51.14       & 59.62         \\
                              & QUAD            & W4A4               & 75.57       & 62.67       & 69.14       & 70.12        & 44.03        & 60.45       & 63.66         \\
                              & QUAD            & W4A4/A8            & 76.50       & 66.30       & 70.76       & 71.21        & 45.99        & 63.48       & 65.71         \\
                              & \;+tuning       & W4A4/A8            & 76.88       & 65.27       & 72.26       & 77.90        & 50.17        & 63.59       & 67.68         \\  \hline
\multirow{5}{*}{Qwen2.5-7B}   & Baseline        & FP16               & 79.98       & 72.69       & 78.90       & 77.31        & 51.11        & 71.71       & 71.95         \\
                              & QuaRot          & W4A4               & 76.55       & 65.59       & 74.59       & 75.00        & 47.95        & 65.44       & 67.52         \\
                              & QUAD            & W4A4               & 78.89       & 68.67       & 75.66       & 74.75        & 48.46        & 67.59       & 69.00         \\
                              & QUAD            & W4A4/A8            & 79.43       & 69.61       & 76.72       & 77.10        & 50.51        & 69.88       & 70.54         \\
                              & \;+tuning       & W4A4/A8            & 79.92       & 68.43       & 77.98       & 82.03        & 55.72        & 70.33       & 72.40         \\  \hline
\end{tabular}
}
\end{table}

\begin{table}[t]
\caption{Generation tasks performance of Qwen-2.5-Instruct models\label{table: generation qwen2.5}}
\centering
\renewcommand\arraystretch{1.5}
\resizebox{0.7\linewidth}{!}{
\begin{tabular}{c|c|c|ccc}
\hline
\textbf{Model}                       & \textbf{Method} & \textbf{Precision} & \textbf{GSM8K}     & \textbf{HumanEval@1} \\ \hline
\multirow{4}{*}{Qwen2.5-3B-Instruct} & Baseline        & FP16               & 47.23              & 48.17              \\ 
                                     & QuaRot          & W4A4               & 37.76              & 20.12              \\ 
                                     & QUAD            & W4A4               & 36.16              & 29.27              \\ 
                                     & QUAD            & W4A4/A8            & 49.96              & 41.46              \\ \hline
\multirow{4}{*}{Qwen2.5-7B-Instruct} & Baseline        & FP16               & 69.75              & 64.02              \\ 
                                     & QuaRot          & W4A4               & 64.44              & 50.00              \\ 
                                     & QUAD            & W4A4               & 68.61              & 51.22              \\ 
                                     & QUAD            & W4A4/A8            & 73.69              & 64.02              \\ \hline
\end{tabular}
}
\end{table}


\begin{table}[t]
\caption{Generation tasks performance of Qwen-2.5 models after parameter-efficient tuning\label{table: peft qwen2.5}}
\centering
\renewcommand\arraystretch{1.5}
\resizebox{0.6\linewidth}{!}{
\begin{tabular}{c|c|c|ccc}
\hline
\textbf{Model}                       & \textbf{Method} & \textbf{Precision} & \textbf{GSM8K}     & \textbf{HumanEval@1} \\ \hline
\multirow{2}{*}{Qwen2.5-3B}          & QLoRA           & W4A16              & 57.13              & 37.20                \\ 
                                     & QUAD            & W4A16              & 57.70              & 38.41                \\ \hline 
\multirow{2}{*}{Qwen2.5-7B}          & QLoRA           & W4A16              & 71.58              & 54.27                \\ 
                                     & QUAD            & W4A16              & 72.25              & 56.10                \\ \hline
\end{tabular}
}
\end{table}

\section{Experiment}

\subsection{Experimental Setup}

\vpara{Models, Datasets, and Baselines.} We evaluate the performance of QUAD on the Llama-2~\cite{llama2-touvron2023}, Llama-3~\cite{llama3-grattafiori2024}, and Qwen-2.5~\cite{qwen2-yang2024} model families across zero-shot and generation tasks using the LM-evaluation-harness framework under default parameter configurations. For zero-shot evaluation, we utilize the benchmark datasets PIQA~\cite{PIQA}, WinoGrande~\cite{winogrande}, HellaSwag~\cite{HellaSwag}, ARC-Easy and ARC-Challenge~\cite{ARC}, and LAMBADA~\cite{LAMBADA}. For generation tasks, we employ GSM8K~\cite{GSM8k} and HumanEval~\cite{HumanEval}. QUAD is compared against the post-training quantization (PTQ) method QuaRot~\cite{quarot-ashkboos2024}.

\vpara{Implementation Details.} QUAD is implemented using PyTorch~\cite{Pytorch} and the Hugging Face Transformers~\cite{huggingface} library. During activation decomposition, the top 64 singular vectors are projected onto an additional dimension. Weight quantization is performed via GPTQ~\cite{gptq-frantar2022}, where the clipping ratio is determined through a linear search over squared error metrics. Activation quantization adopts a round-to-nearest method with a fixed clipping ratio of 0.9, while key-value (KV) caches retain full precision. Symmetric quantization is applied to weights and activations: per-channel quantization for weights and per-token quantization for activations, optimized for efficient GEMM operations. Custom CUDA kernels via Tilelang~\cite{tilelang-tileai2025} are developed for the quantization/dequantization of activations, and leveraging the CUTLASS library~\cite{CUTLASS} to accelerate 4-bit and 8-bit GEMM execution. Calibration datasets consist of 128 samples from the C4 dataset~\cite{C4} for base models and 128 samples from Meta-Math-QA~\cite{MetaMathQA} and Code-Feedback~\cite{CodeFeedback} for instruction-tuned models, each with a sequence length of 2048.

\subsection{Experimental Results}

\vpara{Zero-Shot Task Performance.} Table\ref{table: zero-shot llama-3} and Table \ref{table: zero-shot qwen2.5} compare QUAD and baseline methods on zero-shot accuracy for Llama-3 and Qwen-2 models. Here, "W4A4/A8" denotes 4-bit quantization for U-type layer inputs and 8-bit quantization for D-type layer inputs. QUAD outperforms QuaRot under W4A4 quantization, achieving 93.8\% accuracy for Llama-3-8B and 95.9\% for Qwen-2.5-7B. The hybrid W4A4/A8 configuration strikes an effective precision-performance balance: increasing D-type layer activations to INT8 improves accuracy with only a 35\% increase in GEMM operations compared to full 4-bit quantization. Experiment results for more models (e.g., Llama-2) and baselines (e.g., AWQ, SmoothQuant, GPTQ, and OmniQuant) and efficiency analysis across precision levels are provided in Appendix~\ref{section: additional results}.

\vpara{Generation Task Performance.} Table~\ref{table: generation qwen2.5} evaluates QUAD on generation tasks for the Qwen-2-Instruct family. At W4A4 precision, QUAD surpasses QuaRot by 4.17\% on GSM8K and 1.22\% on HumanEval for Qwen-2.5-7B-Instruct. With W4A4/A8 quantization, QUAD matches the original model's performance on both datasets, demonstrating robustness for generation tasks.

\vpara{Parameter-Efficient Fine-Tuning.} We assess QUAD's fine-tuning capabilities by adapting quantized models on downstream tasks. Zero-shot results for Llama-3 and Qwen-2.5 models fine-tuned on the Alpaca dataset (Table~\ref{table: zero-shot llama-3} and Table~\ref{table: zero-shot qwen2.5}) show that combining QUAD with W4A4/A8 quantization achieves 98.4\% to 100\% of the original model's accuracy. Meanwhile, comparisons with QLoRA~\cite{qlora-dettmers2023} (Table \ref{table: peft qwen2.5}) on Qwen-2.5 models fine-tuned for Meta-Math-QA and Code-Feedback tasks reveal QUAD's superior performance, outperforming QLoRA on GSM8K and HumanEval datasets.

\section{Conclusion}

In this work, we address the challenge of quantizing large language models (LLMs) by proposing QUAD (Quantization with Activation Decomposition). This framework effectively suppresses activation outliers through singular value decomposition (SVD). By decomposing activations into outlier-free components and retaining critical outlier dimensions in full precision, QUAD enables 4-bit quantization of weights and activations while maintaining high accuracy. Our method is compatible with existing rotation-based quantization techniques and introduces a parameter-efficient fine-tuning strategy to narrow the gap between quantized and full-precision models. Experiments on different LLMs demonstrate that QUAD preserves 94–96\% accuracy of the full-precision baseline under W4A4 quantization and reaches 98\% accuracy when combined with W4A4/A8 quantization and fine-tuning. The key contributions of QUAD include (1) a novel SVD-based approach to handle activation outliers and seamless integration with rotation methods and (2) a parameter-efficient fine-tuning mechanism to enhance quantization performance.

\bibliographystyle{unsrt}
\bibliography{main}


\newpage

\appendix

\section{QUAD on the Attention Module\label{section: quad attention}}

\begin{figure}[t]
    \centering
    \includegraphics[width=\linewidth]{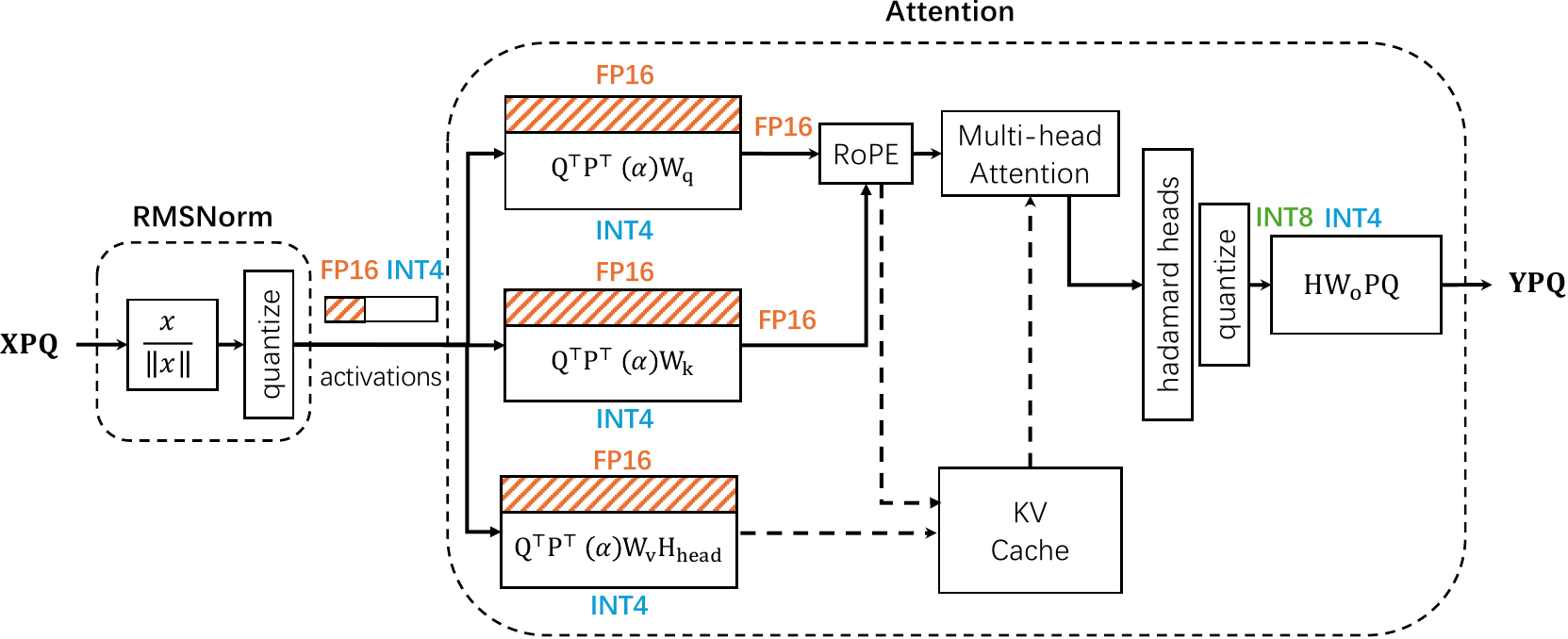} 
    \caption{The Attention module of the transformer layer after applying QUAD transformation.}
    \label{fig: attention quad}
\end{figure}

Figure~\ref{fig: attention quad} shows the result of applying QUAD on the attention module. Based on the original Attention module, we fuse the projection matrix \(P\), the rotation matrix \(Q\), and the scaling factor \((\alpha)\) on the left side of the matrices \(W_q\), \(W_k\), and \(W_v\), and additionally fuse the Hadamard matrix \(H_{\rm head}\) on the right side of \(W_v\).

\begin{table}[t]
\caption{Zero-shot accuracy of LLAMA-2 models.\label{table: zero-shot llama-2}}
\centering
\renewcommand\arraystretch{1.5}
\resizebox{0.75\linewidth}{!}{
\begin{tabular}{c|c|c|ccccccc}
\hline
\textbf{Model}                & \textbf{Method} & \textbf{Precision} & \textbf{PQ} & \textbf{WG} & \textbf{HS} & \textbf{A-e} & \textbf{A-c} & \textbf{LA} & \textbf{Avg.} \\ \hline
\multirow{4}{*}{Llama-2-7B}   & Baseline        & FP16               & 79.11       & 69.06       & 75.99       & 74.54        & 46.33        & 73.84       &   69.81            \\
                              & QuaRot          & W4A4               & 77.80       & 65.04       & 73.23       & 69.82        & 42.66        & 71.53       &    66.68           \\
                              & QUAD            & W4A4               & 77.58       & 68.43       & 73.73       & 71.68        & 41.64        & 73.34       &    67.73           \\
                              & QUAD            & W4A4/A8            & 78.07       & 68.75       & 74.72       & 71.97        & 43.26        & 73.94       &    68.45           \\ \hline
\multirow{4}{*}{Llama-2-13B}  & Baseline        & FP16               & 80.52       & 72.38       & 79.38       & 77.44        & 49.15        & 76.77       &    72.61           \\
                              & QuaRot          & W4A4               & 77.69       & 69.22       & 76.02       & 74.54        & 47.78        & 74.48       &    69.96           \\
                              & QUAD            & W4A4               & 79.00       & 71.11       & 77.45       & 75.08        & 46.76        & 75.86       &    70.88           \\
                              & QUAD            & W4A4/A8            & 79.71       & 72.30       & 78.11       & 75.63        & 47.87        & 76.56       &    71.70      \\ \hline
\end{tabular}
}
\end{table}

\section{Hadamard Transform\label{section: hadamard}}






A Hadamard matrix is an orthogonal matrix whose entries belong to \(\{+1, -1\}\). A Walsh-Hadamard matrix is a square matrix of size \(d = 2^n\), defined recursively as:

\[
    H_2 = \frac{1}{\sqrt{2}} \begin{bmatrix}
        1 & 1 \\
        1 & -1
    \end{bmatrix}, \quad \text{and} \quad H_{2^n} = H_2 \otimes H_{2^{n-1}},
\]

where \(\otimes\) represents the Kronecker product. These definitions underpin the Walsh-Hadamard transform, which efficiently computes the matrix-vector product \(Hx\) in \(O(d \log_2(d))\) operations.

Furthermore, prior work in QuaRot~\cite{quarot-ashkboos2024} and QUIP~\cite{quip-chee2023} demonstrates that applying the Hadamard transform to tensors reduces the incoherence of weight matrices and activations. This reduction simplifies the difficulty of quantization. Specifically, a weight matrix \(W\) is considered \(\mu\)-incoherent if \(\max(W) \leq \mu \cdot \|W\|_F \cdot \frac{1}{\sqrt{mn}}\), where \(\max(W)\) denotes the element-wise maximum of the matrix, and \(mn\) represents the total number of elements in \(W\). In this work, we follow the Hadamard transform used in existing work~\cite{quarot-ashkboos2024, spinquant-liu2024, ostquant-hu2025} to smooth the weights and activations, thus reducing the quantization difficulty.

\section{Proofs\label{section: proofs}}

\subsection{Proof about projection matrix\label{section: proof of PPt = I}}

\begin{proof}
    \begin{align*}
        P P^\top & = \left( U_{1:r}, I - \sum_{i=1}^{r} U_i U_i^\top \right) \left( U_{1:r}, I - \sum_{i=1}^{r} U_i U_i^\top \right)^\top \\
                 & = \left( U_{1:r}, I - \sum_{i=1}^{r} U_i U_i^\top \right) \left(\begin{matrix}
                     U_{1:r}^\top, \\
                     I - \sum_{i=1}^{r} U_i U_i^\top
                 \end{matrix} \right) \\
                 & = \left( U_{1:r} U_{1:r}^\top + (I - \sum_{i=1}^{r} U_i U_i^\top) (I - \sum_{i=1}^{r} U_i U_i^\top) \right) \\
                 & = \left( \sum_{i=1}^{r} U_i U_i^\top + I - 2\sum_{i=1}^{r} U_i U_i^\top I + \sum_{i=1}^{r} U_i U_i^\top \right) \\
                 & = I
    \end{align*}
\end{proof}

\subsection{Proof of Proposition~\ref{prop: quant error}}

\textbf{Proposition.} \textit{The quantization error can be decomposed as follows}:
\[
    E(X, W) \leq \Vert X \Vert_F \Vert W - Q(W) \Vert_F + \Vert X - Q(X) \Vert_F \Vert Q(W) \Vert_F.
\]

\begin{proof}

\begin{align*}
    E(X, W) & = \Vert XW - Q(X)Q(W)\Vert_F \\
    & = \Vert XW - XQ(W) + XQ(W) - Q(X)Q(W) \Vert \\
    & \leq \Vert X (W - Q(W)) \Vert_F + \Vert (X - Q(X))Q(W) \Vert_F \\
    & \leq \Vert X \Vert_F \Vert W - Q(W) \Vert_F + \Vert X - Q(X) \Vert_F \Vert Q(W) \Vert_F.
\end{align*}

\end{proof}

\subsection{Proof of Proposition~\ref{prop: rtn bound}}

\textbf{Proposition.} \textit{For the round-to-nearest (RTN) quantization method \(Q\), if the tensor \(X\) follows a normal distribution, we have}:
    \[
        \mathbb{E} \left[ \Vert X - Q(X) \Vert_F \right] \leq \frac{\sqrt{\log (size(X) \pi)}}{q_{\text{max}}} \mathbb{E} \left[ \Vert X \Vert_F \right],
    \]

The proof of proposition~\ref{prop: rtn bound} can be found in Section A.2 of SVDQuant~\cite{svdqunat-li2024}.

\subsection{Proof of Proposition~\ref{prop: approx}}

\textbf{Proposition.} \textit{Suppose the gradients of the weights \(W\) and \(W_r\) are \(\nabla W\) and \(\nabla W_r\), then we have}:
\[
    X \nabla W = XX^\top \nabla Y \approx X_r \nabla W_r.
\]


\begin{proof}
    Consider the singular value decomposition (SVD) of the matrix $X^\top$, which is given as $X^\top = U \Sigma V^\top$. From this, we can derive the SVDs of $XX^\top$ and $X^\top X$:
    \[
    \text{SVD}(XX^\top) = V \Sigma^2 V^\top, \text{SVD}(X^\top X) = U \Sigma^2 U^\top.
    \]
    According to the Eckart-Young-Mirsky theorem, a matrix can be approximated optimally using its singular value decomposition. Specifically, for any matrix $D$, the best rank-$r$ approximation is given by $U_{1:r} \Sigma_{1:r} V_{1:r}^\top$, where $\text{SVD}(D) = U \Sigma V^\top$. Therefore, the optimal rank-$r$ approximation of $XX^\top$ is \(V_{1:r} \Sigma^2_{1:r} V_{1:r}^\top\).
    
    Let $X_r = X U_{1:r}$. Then, we compute $X_r X_r^\top$ as follows:
    \begin{align*}
        X_r X_r^\top &= X U_{1:r} U_{1:r}^\top X^\top \\
                     &= V \Sigma U^\top U_{1:r} U_{1:r}^\top U \Sigma V^\top \\
                     &= V \Sigma \begin{pmatrix}
                            I_r & 0 \\
                            0   & 0
                        \end{pmatrix} \Sigma V^\top \\
                     &= V_{1:r} \Sigma^2_{1:r} V_{1:r}^\top.
    \end{align*}
    Thus, $X_rX_r^\top$ is the optimal rank-$r$ approximation of $XX^\top$.
\end{proof}

\begin{table}[t]
\caption{Zero-shot accuracy and computation efficiency of LLAMA-3-8B in different precision, where Avg. denotes the average accuracy and Pct. Denotes the percentage of INT4 GEMM (INT8 GEMM is used for the rest).\label{table: Avg. and Pct.}}
\centering
\renewcommand\arraystretch{1.5}
\resizebox{0.6\linewidth}{!}{
\begin{tabular}{l|ccc}
\hline
                & \textbf{QuaRot-W4A4} & \textbf{QuaRot-W4A8} & \textbf{QUAD-W4A4/A8}   \\ \hline
\textbf{Avg.}   & 65.43                & 72.30                &  70.71                  \\ 
\textbf{Pct.}   & 100\%                & 0\%                  &  65.4\%                 \\ \hline 
                & \textbf{QUAD-W4A4}   & \textbf{QUAD-W4A8}   & \textbf{QUAD-W4A4/A8-tuning}   \\ \hline
\textbf{Avg.}   & 68.76                & 72.33                &  72.14                  \\ 
\textbf{Pct.}   & 100\%                & 0\%                  &  65.4\%                 \\ \hline 
\end{tabular}
}
\end{table}

\begin{figure}[t]
    \centering
    \includegraphics[width=0.6\linewidth]{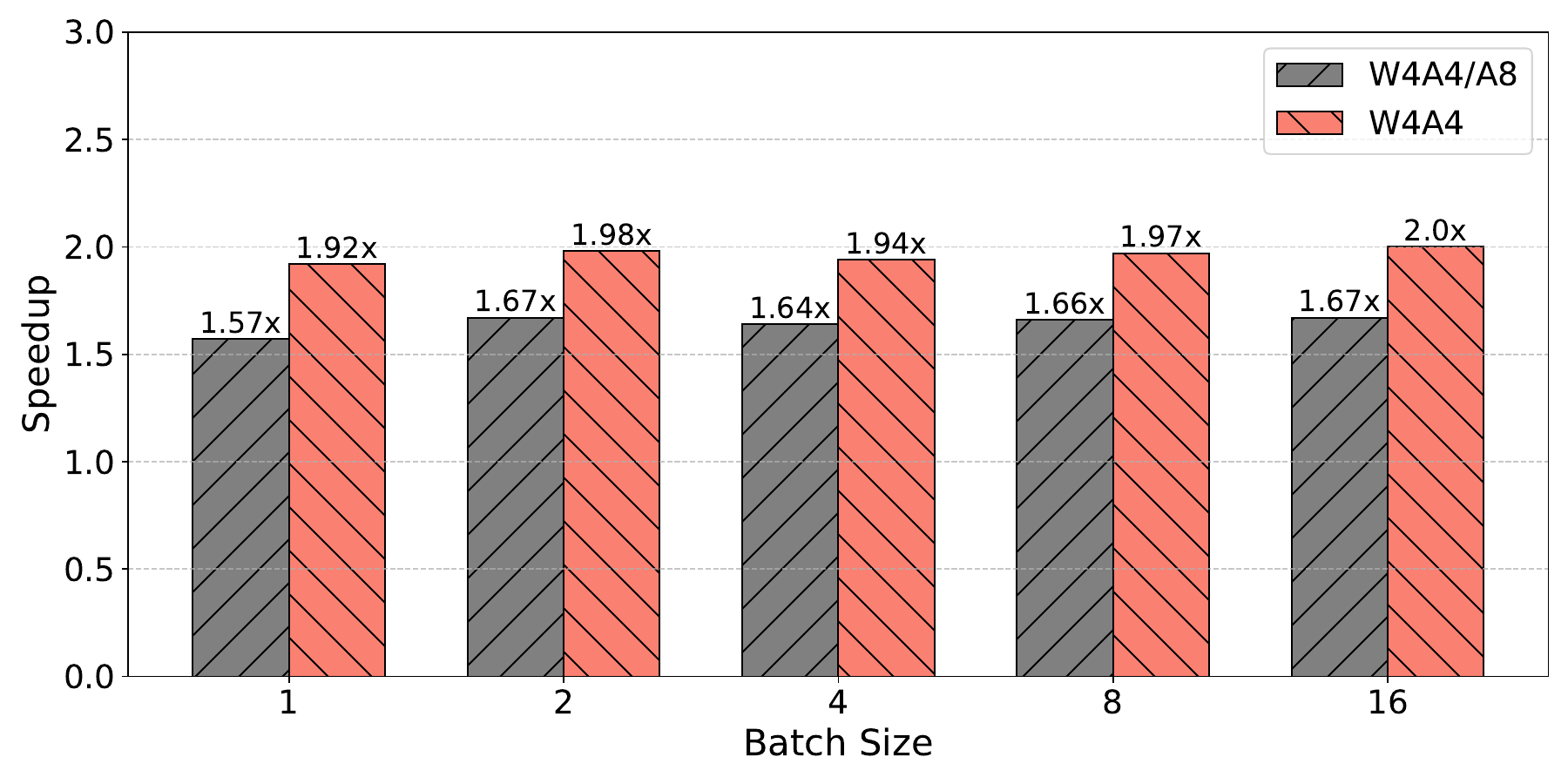} 
    \caption{Speedup of QUAD on Llama-3-8B under sequence length 2048.}
    \label{fig: speedup}
\end{figure}

\section{Additional Experiment Results\label{section: additional results}}



Table~\ref{table: zero-shot llama-2} presents the experimental results for QUAD and QuaRot on the Llama-2 model. The results indicate that QUAD outperforms QuaRot, underscoring its effectiveness in achieving higher accuracy in zero-shot tasks. Furthermore, we compared QUAD with additional baselines on Llama-2-7B, Llama-2-13B, and Llama-3-8B, and the corresponding experimental results are detailed in Table~\ref{table: zero-shot llama-2 and llama-3}.

Table~\ref{table: Avg. and Pct.} shows the accuracy of QUAD and QuaRot under the W4A4, W4A8, and W4A4/A8 quantization schemas, along with the percentage of INT4 and INT8 GEMM operations utilized. These findings highlight the superior balance between efficiency and performance achieved by the W4A4/A8 quantization approach relative to W4A4 and W4A8 quantization alone. Additionally, we assessed the prefill speed of Llama-3-8B with W4A4 and W4A4/A8 quantization schemas, setting the sequence length to 2048. The experimental results, illustrated in Figure~\ref{fig: speedup}, reveal approximately \(1.6\times\) and \(2.0\times\) speedups compared to the baseline when using W4A4/A8 and W4A4 quantization schemas, respectively.

\section{Eliminating Online Hadamard Transform\label{section: low-rank branch}}

\begin{figure}[t]
    \centering
    \includegraphics[width=0.6\linewidth]{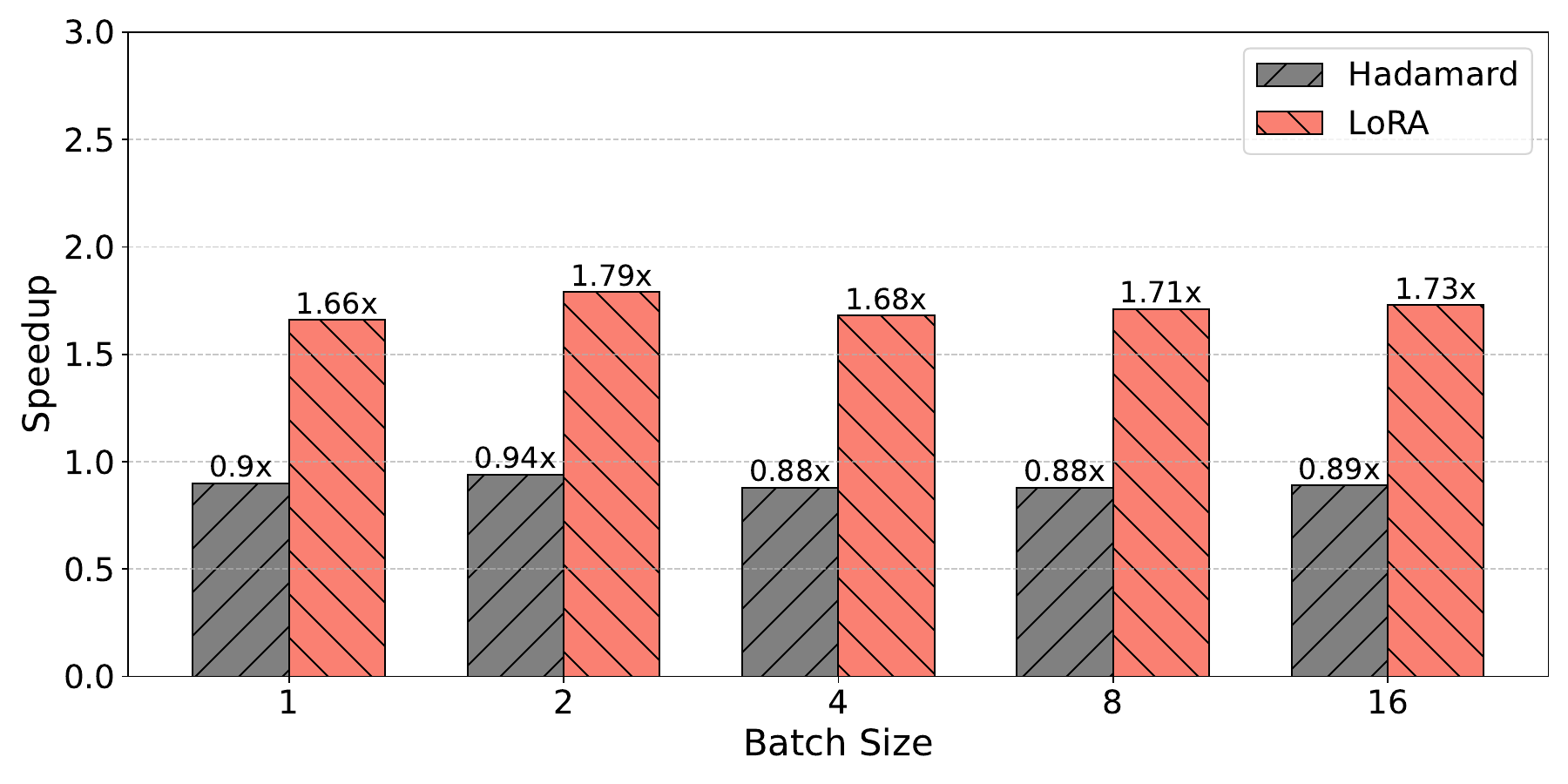} 
    \caption{Speedup of QUAD-Hadamard/LoRA on Qwen-2.5-7B under sequence length 2048 and W4A4/A8 quantization schema.}
    \label{fig: speedup-2}
\end{figure}

For the Attention module, both QUAD and existing approaches smooth the scale-dot-product-attention (SDPA) output using the online Hadamard transform. This approach is efficient for models where the number of attention heads is a power of 2, such as Llama-2-7B and Llama-3-8B. However, for models that do not meet this condition, such as the Qwen-2.5-7B and Llama-2-13B, it introduces substantial latency. As illustrated in Figure~\ref{fig: speedup-2}, under the W4A4/A8 quantization schema, employing the online Hadamard transform results in slower inference speeds compared to the baseline using FP16 precision. To address this issue, inspired by SVDQuant, we experimented with replacing the online Hadamard transform with full-precision low-rank branches. Specifically, for the Hadamard transform, we have:

\begin{equation*}
    F_{had}(X, W) = Q(\text{Hadamard}(X)) Q(H W),
\end{equation*}

where \( Q \) denotes quantization. In the model, we replaced \( F_{had} \) with \( F_{LoRA} \), which corresponds to the following expression:

\begin{equation*}
    F_{LoRA}(X,W',L,R) = Q(X s^{-1})Q(sW') + Xs^{-1}LR,
\end{equation*}

where \( s_i = \max(|X_i|)^{0.25} \), \( L = U \), \( R = \Sigma V^\top \), \( W' = W - LR \), and \( U \Sigma V = \text{SVD}(sW) \). As shown in Figure~\ref{fig: speedup-2}, replacing the Hadamard transform with LoRA yields a significant improvement in the model's inference speed, achieving approximately \( 1.7 \times \) speedup relative to the baseline. Additionally, we compared the performance of the model under the two approaches, and the experimental results are presented in Table~\ref{table: Hadamard vs LoRA}. These results demonstrate that the use of low-rank branching can achieve competitive performance with the Hadamard transform within the W4A4/A8 quantization schema.

\begin{table}[t]
\caption{Comparison of Hadamard transform and low-rank branch on LLAMA and Qwen models with W4A4/A8 quantization schema.\label{table: Hadamard vs LoRA}}
\centering
\renewcommand\arraystretch{1.5}
\resizebox{0.85\linewidth}{!}{
\begin{tabular}{c|c|ccccccc}
\hline
\textbf{Model}               & \textbf{Method} & \textbf{PQ} & \textbf{WG} & \textbf{HS} & \textbf{A-e} & \textbf{A-c} & \textbf{LA} & \textbf{Avg.} \\ \hline
\multirow{2}{*}{Llama-2-7B}  & Hadamard        & 78.07       & 68.75       & 74.72       & 71.97        & 43.26        & 73.94       & 68.45         \\
                             & LoRA            & 79.27       & 68.59       & 74.27       & 72.90        & 43.34        & 72.44       & 68.47         \\ \hline
\multirow{2}{*}{Llama-2-13B} & Hadamard        & 79.71       & 72.30       & 78.11       & 75.63        & 47.87        & 76.56       & 71.70         \\
                             & LoRA            & 79.27       & 71.43       & 77.81       & 76.56        & 48.12        & 77.78       & 71.83         \\ \hline
\multirow{2}{*}{Llama-3-8B}  & Hadamard        & 80.09       & 71.35       & 77.12       & 73.82        & 47.53        & 74.35       & 70.71         \\
                             & LoRA            & 78.67       & 72.06       & 76.73       & 76.77        & 48.98        & 74.36       & 71.26         \\ \hline
\multirow{2}{*}{Qwen-2.5-1.5B} & Hadamard      & 73.88       & 60.22       & 64.38       & 70.54        & 42.83        & 59.31       & 61.86         \\
                               & LoRA          & 72.85       & 61.48       & 64.06       & 70.92        & 41.64        & 57.27       & 61.37         \\ \hline
\multirow{2}{*}{Qwen-2.5-3B} & Hadamard        & 76.50       & 66.30       & 70.76       & 71.21        & 45.99        & 63.48       & 65.71         \\
                             & LoRA            & 76.44       & 63.85       & 70.85       & 73.44        & 46.42        & 62.78       & 65.63         \\ \hline
\multirow{2}{*}{Qwen-2.5-7B} & Hadamard        & 79.43       & 69.61       & 76.72       & 77.10        & 50.51        & 69.88       & 70.54         \\
                             & LoRA            & 77.91       & 69.61       & 76.44       & 76.01        & 49.74        & 68.97       & 69.78         \\ \hline
\end{tabular}
}
\end{table}

\begin{table}[t]
\caption{Complete experiment results of LLAMA-2 and LLAMA-3 models.\label{table: zero-shot llama-2 and llama-3}}
\centering
\renewcommand\arraystretch{1.2}
\resizebox{\linewidth}{!}{
\begin{tabular}{c|c|c|cccccccccc}
\hline
\textbf{Model}               & \textbf{\begin{tabular}[c]{@{}c@{}}\#Bits\\ W-A-KV\end{tabular}} & \textbf{Method} & \textbf{A-c} & \textbf{A-e} & \textbf{BQ} & \textbf{HS} & \textbf{LA} & \textbf{OQ} & \textbf{PQ} & \textbf{SQ} & \textbf{WG} & \textbf{Avg.} \\ \hline
\multirow{19}{*}{Llama-2-7B} & 16-16-16                                                         & Baseline        & 46.33        & 74.54        & 77.71       & 75.99       & 73.84       & 44.20       & 79.11       & 46.11       & 69.06       & 65.21         \\ \cline{2-13} 
                             & \multirow{7}{*}{4-16-16}                                         & RTN             & 42.15        & 67.59        & 73.06       & 72.34       & 67.18       & 41.80       & 76.50       & 44.11       & 66.69       & 61.27         \\
                             &                                                                  & SmoothQuant     & 39.59        & 65.19        & 69.82       & 68.84       & 62.27       & 40.20       & 75.95       & 44.17       & 63.85       & 58.88         \\
                             &                                                                  & GPTQ            & 42.49        & 69.53        & 61.31       & 73.83       & 67.61       & 42.40       & 77.64       & 44.52       & 68.43       & 60.86         \\
                             &                                                                  & OmniQuant       & 42.49        & 71.00        & 74.34       & 73.85       & 70.70       & 44.20       & 78.40       & 44.93       & 68.82       & 63.19         \\
                             &                                                                  & AWQ             & 44.11        & 70.75        & 78.07       & 74.98       & 70.68       & 43.80       & 78.13       & 45.14       & 69.38       & 63.89         \\
                             &                                                                  & QuaRot          & 43.94        & 73.15        & 76.97       & 74.87       & 73.06       & 44.00       & 78.24       & 45.09       & 69.38       & 64.30         \\
                             &                                                                  & QUAD            & 44.11        & 73.19        & 76.51       & 75.30       & 74.27       & 43.80       & 78.62       & 46.06       & 70.01       & 64.65         \\ \cline{2-13} 
                             & \multirow{5}{*}{4-4-16}                                          & RTN             & 25.34        & 28.03        & 50.52       & 27.71       & 1.01        & 26.20       & 50.82       & 33.93       & 48.38       & 32.44         \\
                             &                                                                  & SmoothQuant     & 28.33        & 26.39        & 49.39       & 27.28       & 1.18        & 23.40       & 48.80       & 33.62       & 50.75       & 32.13         \\
                             &                                                                  & GPTQ            & 24.40        & 28.70        & 51.62       & 28.66       & 1.36        & 24.60       & 51.14       & 34.49       & 49.49       & 32.72         \\
                             &                                                                  & QuaRot          & 42.32        & 69.65        & 74.77       & 72.91       & 70.81       & 39.80       & 77.20       & 43.55       & 65.82       & 61.87         \\
                             &                                                                  & QUAD            & 41.64        & 71.68        & 74.22       & 73.73       & 73.34       & 43.00       & 77.58       & 45.70       & 68.43       & 63.26         \\ \cline{2-13} 
                             & \multirow{6}{*}{4-4-4}                                           & RTN             & 27.22        & 27.06        & 50.83       & 27.34       & 0.93        & 25.80       & 49.51       & 34.85       & 50.51       & 32.67         \\
                             &                                                                  & SmoothQuant     & 26.37        & 25.63        & 47.71       & 27.05       & 1.11        & 26.40       & 51.90       & 34.49       & 48.38       & 32.12         \\
                             &                                                                  & GPTQ            & 26.96        & 27.65        & 52.84       & 28.83       & 1.63        & 29.20       & 49.62       & 35.11       & 49.80       & 33.52         \\
                             &                                                                  & OmniQuant       & 31.40        & 53.75        & 63.79       & 55.06       & 35.63       & 34.40       & 66.59       & 40.28       & 54.70       & 48.40         \\
                             &                                                                  & QuaRot          & 41.43        & 69.32        & 74.19       & 72.50       & 70.66       & 39.80       & 77.42       & 43.35       & 64.64       & 61.48         \\
                             &                                                                  & QUAD            & 41.64        & 71.68        & 74.22       & 73.73       & 73.34       & 43.00       & 77.58       & 45.70       & 68.43       & 63.26         \\ \hline
\multirow{19}{*}{Llama-2-13B}& 16-16-16                                                         & Baseline        & 49.15        & 77.53        & 80.58       & 79.39       & 76.62       & 45.20       & 80.63       & 47.49       & 71.90       & 67.61         \\ \cline{2-13} 
                             & \multirow{7}{*}{4-16-16}                                         & RTN             & 42.92        & 66.54        & 71.38       & 66.62       & 68.99       & 39.40       & 76.93       & 44.06       & 65.35       & 60.24         \\
                             &                                                                  & SmoothQuant     & 46.25        & 70.45        & 74.92       & 69.16       & 70.49       & 39.80       & 77.86       & 45.14       & 64.17       & 62.03         \\
                             &                                                                  & GPTQ            & 49.63        & 73.95        & 74.83       & 73.77       & 73.20       & 42.40       & 78.51       & 45.50       & 70.64       & 64.71         \\
                             &                                                                  & OmniQuant       & 48.29        & 75.42        & 77.92       & 77.80       & 75.59       & 45.20       & 80.41       & 46.62       & 70.17       & 66.38         \\
                             &                                                                  & AWQ             & 48.63        & 78.16        & 78.81       & 78.48       & 75.20       & 45.00       & 79.54       & 46.21       & 72.45       & 66.94         \\
                             &                                                                  & QuaRot          & 49.15        & 76.26        & 80.46       & 78.17       & 76.50       & 45.40       & 80.03       & 45.50       & 71.11       & 66.95         \\
                             &                                                                  & QUAD            & 47.61        & 75.42        & 79.24       & 78.87       & 77.28       & 45.40       & 79.65       & 45.96       & 72.30       & 66.86         \\ \cline{2-13} 
                             & \multirow{5}{*}{4-4-16}                                          & RTN             & 27.99        & 26.81        & 38.50       & 26.08       & 0.00        & 23.60       & 48.20       & 34.90       & 51.62       & 30.86         \\
                             &                                                                  & SmoothQuant     & 24.49        & 35.06        & 47.98       & 30.87       & 3.67        & 26.20       & 55.01       & 35.31       & 49.72       & 34.26         \\
                             &                                                                  & GPTQ            & 27.82        & 26.77        & 37.92       & 25.67       & 0.00        & 21.80       & 47.77       & 35.11       & 48.15       & 30.11         \\
                             &                                                                  & QuaRot          & 46.42        & 73.86        & 78.10       & 75.68       & 74.31       & 43.00       & 79.05       & 44.37       & 71.35       & 65.13         \\
                             &                                                                  & QUAD            & 46.76        & 75.08        & 78.69       & 77.45       & 75.86       & 43.40       & 79.00       & 45.14       & 71.11       & 65.83         \\ \cline{2-13} 
                             & \multirow{6}{*}{4-4-4}                                           & RTN             & 27.82        & 26.52        & 38.38       & 26.27       & 0.02        & 26.00       & 49.78       & 34.39       & 49.17       & 30.93         \\
                             &                                                                  & SmoothQuant     & 24.49        & 33.00        & 45.84       & 30.70       & 2.70        & 23.80       & 53.81       & 34.80       & 51.07       & 33.36         \\
                             &                                                                  & GPTQ            & 27.90        & 26.39        & 37.95       & 26.16       & 0.00        & 27.00       & 48.26       & 34.39       & 50.43       & 30.94         \\
                             &                                                                  & OmniQuant       & 32.85        & 55.13        & 64.34       & 60.13       & 42.85       & 33.40       & 68.17       & 39.76       & 56.51       & 50.35         \\
                             &                                                                  & QuaRot          & 47.27        & 73.91        & 78.41       & 75.33       & 73.53       & 43.80       & 79.27       & 45.85       & 69.06       & 65.16         \\
                             &                                                                  & QUAD            & 46.50        & 74.96        & 78.47       & 77.32       & 76.01       & 44.20       & 80.03       & 45.09       & 71.27       & 65.98         \\ \hline
\multirow{19}{*}{Llama-3-8B} & 16-16-16                                                         & Baseline        & 53.50        & 77.74        & 81.10       & 79.18       & 75.74       & 44.80       & 80.63       & 47.08       & 73.01       & 68.09         \\ \cline{2-13} 
                             & \multirow{7}{*}{4-16-16}                                         & RTN             & 48.98        & 73.23        & 72.75       & 75.90       & 63.85       & 43.20       & 78.40       & 43.81       & 73.16       & 63.70         \\
                             &                                                                  & SmoothQuant     & 47.44        & 72.35        & 72.11       & 74.92       & 62.41       & 43.00       & 77.69       & 43.91       & 71.27       & 62.79         \\
                             &                                                                  & GPTQ            & 49.74        & 72.52        & 71.28       & 68.34       & 46.69       & 43.60       & 78.78       & 46.47       & 71.82       & 61.03         \\
                             &                                                                  & OmniQuant       & 50.09        & 74.54        & 79.51       & 76.92       & 70.31       & 43.80       & 79.54       & 44.52       & 71.74       & 65.66         \\
                             &                                                                  & AWQ             & 52.22        & 76.68        & 80.31       & 77.51       & 74.81       & 44.20       & 79.60       & 46.26       & 71.67       & 67.03         \\
                             &                                                                  & QuaRot          & 51.88        & 77.53        & 79.60       & 77.87       & 74.11       & 44.40       & 80.14       & 46.37       & 73.56       & 67.27         \\
                             &                                                                  & QUAD            & 50.77        & 75.34        & 80.58       & 77.93       & 75.63       & 45.00       & 80.63       & 46.26       & 73.80       & 67.33         \\ \cline{2-13} 
                             & \multirow{5}{*}{4-4-16}                                          & RTN             & 23.72        & 30.89        & 46.30       & 31.26       & 3.03        & 27.60       & 52.72       & 35.26       & 50.04       & 33.42         \\
                             &                                                                  & SmoothQuant     & 23.29        & 28.28        & 48.93       & 29.19       & 1.57        & 28.60       & 54.46       & 33.37       & 49.64       & 33.04         \\
                             &                                                                  & GPTQ            & 23.46        & 32.07        & 43.79       & 30.10       & 2.41        & 28.00       & 53.97       & 34.14       & 48.86       & 32.98         \\
                             &                                                                  & QuaRot          & 42.66        & 67.26        & 73.73       & 73.60       & 67.42       & 43.00       & 76.61       & 45.04       & 65.90       & 61.69         \\
                             &                                                                  & QUAD            & 46.50        & 73.74        & 76.02       & 75.22       & 71.18       & 43.00       & 77.48       & 44.42       & 68.43       & 64.00         \\ \cline{2-13} 
                             & \multirow{6}{*}{4-4-4}                                           & RTN             & 23.72        & 30.56        & 46.18       & 29.83       & 2.70        & 28.60       & 52.45       & 34.39       & 50.20       & 33.18         \\
                             &                                                                  & SmoothQuant     & 23.55        & 28.96        & 48.84       & 28.90       & 1.44        & 29.40       & 51.09       & 34.14       & 50.36       & 32.96         \\
                             &                                                                  & GPTQ            & 23.38        & 32.74        & 44.34       & 29.72       & 2.39        & 29.80       & 54.95       & 34.75       & 51.30       & 33.71         \\
                             &                                                                  & OmniQuant       & 22.87        & 30.35        & 41.53       & 31.11       & 1.86        & 25.40       & 53.37       & 34.08       & 50.43       & 32.33         \\
                             &                                                                  & QuaRot          & 42.83        & 67.42        & 73.21       & 72.66       & 66.93       & 42.20       & 75.73       & 45.19       & 66.22       & 61.38         \\
                             &                                                                  & QUAD            & 45.90        & 72.10        & 73.98       & 74.90       & 69.82       & 43.40       & 78.45       & 44.22       & 71.43       & 63.80         \\ \hline

\end{tabular}
}
\end{table}


\end{document}